\documentclass[10pt,twocolumn,letterpaper]{article}

\usepackage[pagenumbers]{cvpr} 

\usepackage{graphicx}
\usepackage{amsmath}
\usepackage{amssymb}
\usepackage{booktabs}
\usepackage{comment}
\usepackage{svg}
\usepackage{bbding}
\usepackage{makecell}
\usepackage{multirow}
\usepackage{tabularx}
\usepackage[labelsep=period]{caption}
\usepackage{tikz}   
\usepackage{indentfirst} 
\usepackage{colortbl} 

\usepackage[pagebackref,breaklinks,colorlinks,citecolor=cvprblue]{hyperref}
\usepackage{url}

\definecolor{ao}{rgb}{0.0, 0.0, 1.0}
\definecolor{airforceblue}{rgb}{0.36, 0.54, 0.66}
\definecolor{ceruleanblue}{rgb}{0.16, 0.32, 0.75}
\definecolor{cerulean}{rgb}{0.0, 0.48, 0.65}
\definecolor{celestialblue}{rgb}{0.29, 0.59, 0.82}
\definecolor{azure(colorwheel)}{rgb}{0.0, 0.5, 1.0}
\definecolor{babyblue}{rgb}{0.54, 0.81, 0.94}
\definecolor{babyblueeyes}{rgb}{0.63, 0.79, 0.95}
\definecolor{ballblue}{rgb}{0.13, 0.67, 0.8}

\definecolor{asparagus}{rgb}{0.53, 0.66, 0.42}
\definecolor{ao(english)}{rgb}{0.0, 0.5, 0.0}
\definecolor{applegreen}{rgb}{0.55, 0.71, 0.0}
\definecolor{armygreen}{rgb}{0.29, 0.33, 0.13}
\definecolor{gray-asparagus}{rgb}{0.27, 0.35, 0.27}
\definecolor{green(ryb)}{rgb}{0.4, 0.69, 0.2}

\definecolor{amethyst}{rgb}{0.6, 0.4, 0.8}
\definecolor{antiquefuchsia}{rgb}{0.57, 0.36, 0.51}
\definecolor{blue-violet}{rgb}{0.54, 0.17, 0.89}
\definecolor{brightlavender}{rgb}{0.75, 0.58, 0.89}
\definecolor{brightube}{rgb}{0.82, 0.62, 0.91}
\definecolor{brilliantlavender}{rgb}{0.96, 0.73, 1.0}

\definecolor{amber}{rgb}{1.0, 0.75, 0.0}
\definecolor{amber(sae/ece)}{rgb}{1.0, 0.49, 0.0}
\definecolor{atomictangerine}{rgb}{1.0, 0.6, 0.4}
\definecolor{burntorange}{rgb}{0.8, 0.33, 0.0}
\definecolor{burntsienna}{rgb}{0.91, 0.45, 0.32}
\definecolor{cadmiumorange}{rgb}{0.93, 0.53, 0.18}
\definecolor{carrotorange}{rgb}{0.93, 0.57, 0.13}
\definecolor{chocolate(web)}{rgb}{0.82, 0.41, 0.12}
\definecolor{chromeyellow}{rgb}{1.0, 0.65, 0.0}
\definecolor{darkorange}{rgb}{1.0, 0.55, 0.0}
\definecolor{darktangerine}{rgb}{1.0, 0.66, 0.07}
\definecolor{deepcarrotorange}{rgb}{0.91, 0.41, 0.17}
\definecolor{deepsaffron}{rgb}{1.0, 0.6, 0.2}
\definecolor{fulvous}{rgb}{0.86, 0.52, 0.0}

\newcommand{\thename}{MVHumanNet}

\definecolor{cvprblue}{rgb}{0.21,0.49,0.74}
\definecolor{darkgreen}{rgb}{0.0, 0.5, 0.0}
\definecolor{darkred}{rgb}{0.55, 0.0, 0.0}   
\definecolor{deeppeach}{rgb}{1.0, 0.8, 0.64}  
\definecolor{lightgray}{rgb}{0.83, 0.83, 0.83}  

\usepackage{pifont}
\usepackage[marginal]{footmisc}

\usepackage{multirow}
\usepackage{makecell}

\usepackage[capitalize]{cleveref}
\crefname{section}{Sec.}{Secs.}
\Crefname{section}{Section}{Sections}
\Crefname{table}{Table}{Tables}
\crefname{table}{Tab.}{Tabs.}

\title{ \vspace{-4mm} MVHumanNet: A Large-scale Dataset of Multi-view Daily \\ Dressing  Human Captures }

\author{Zhangyang Xiong$^{1,2}$$^{\#}$ \quad Chenghong Li$^{1,2}$$^{\#}$ \quad Kenkun Liu$^{2}$$^{\#}$  \quad Hongjie Liao$^{2}$ \quad Jianqiao Hu$^{2}$ \\ 
\quad Junyi Zhu$^{2}$\quad Shuliang Ning$^{1,2}$\quad Lingteng Qiu$^{2}$  \quad Chongjie Wang$^{2}$\quad Shijie Wang$^{2}$\quad \\
Shuguang Cui$^{2,1}$\quad Xiaoguang Han$^{2,1*}$ \vspace{2pt}\\
\small{$^{\#}$equal contribution} \qquad \small{$^{*}$corresponding author} \vspace{2pt}\\
{\normalsize $^1$FNii, CUHKSZ}\qquad
{\normalsize $^2$SSE, CUHKSZ} 
}

\begin{document}

\vspace{-6mm}
\twocolumn[{%
\renewcommand\twocolumn[1][]{#1}%
\maketitle
\vspace{-8mm}
\begin{center}
    \centering
    \captionsetup{type=figure}
    \includegraphics[width=1\textwidth]{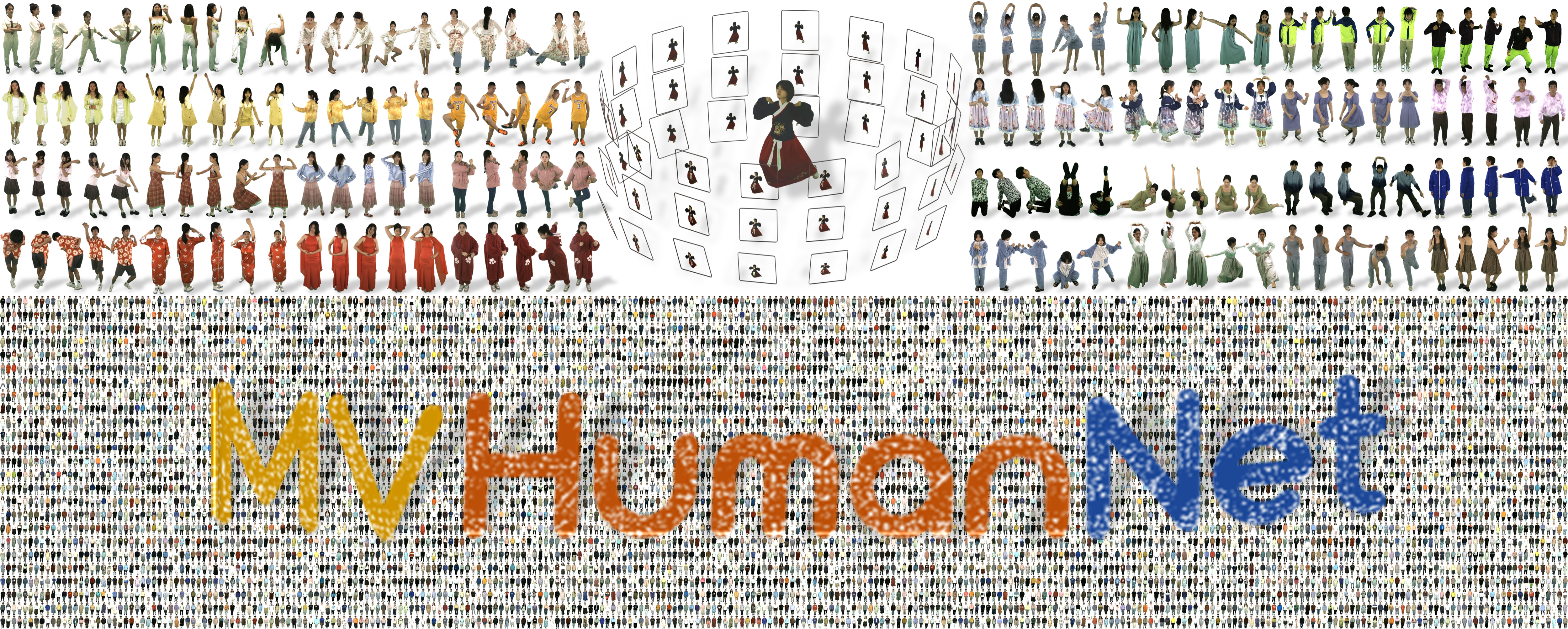}
    \captionsetup{skip=4pt} 
    \captionof{figure}{
    We introduce \textbf{MVHumanNet}, a large-scale dataset of \textit{multi-view human images} with unprecedented scale in human subjects, daily outfits, motion sequences and frames. 
    \textbf{Top left and right}: Examples of multi-view poses featuring different human identities with various daily dressing in our dataset. 
    \textbf{Top middle}: Our multi-view capture system includes 48 cameras of 12MP resolution. 
    \textbf{Bottom}: Comprehensive visualization of all 9000 outfits in our MVHumanNet.
    \label{fig:teaser}}
\end{center}%
}] 


\vspace{-6mm}
\maketitle

\begin{abstract}
\vspace{-4mm}
In this era, the success of large language models and text-to-image models can be attributed to the driving force of large-scale datasets. However, in the realm of 3D vision, while remarkable progress has been made with models trained on large-scale synthetic and real-captured object data like Objaverse and MVImgNet, a similar level of progress has not been observed in the domain of human-centric tasks partially due to the lack of a large-scale human dataset. Existing datasets of high-fidelity 3D human capture continue to be mid-sized due to the significant challenges in acquiring large-scale high-quality 3D human data. To bridge this gap, we present \textbf{MVHumanNet}, a dataset that comprises multi-view human action sequences of \textbf{4,500} human identities. The primary focus of our work is on collecting human data that features a large number of diverse identities and everyday clothing using a multi-view human capture system, which facilitates easily scalable data collection. Our dataset contains \textbf{9,000} daily outfits, \textbf{60,000} motion sequences and \textbf{645 million} frames with extensive annotations, including human masks, camera parameters , 2D and 3D keypoints, SMPL/SMPLX parameters, and corresponding textual descriptions. To explore the potential of MVHumanNet in various 2D and 3D visual tasks, we conducted pilot studies on view-consistent action recognition, human NeRF reconstruction, text-driven view-unconstrained human image generation, as well as 2D view-unconstrained human image and 3D avatar generation.  Extensive experiments demonstrate the performance improvements and effective applications enabled by the scale provided by MVHumanNet. As the current largest-scale 3D human dataset, we hope that the release of MVHumanNet data with annotations will foster further innovations in the domain of 3D human-centric tasks at scale.
\end{abstract}

\section{Introduction}
\label{sec:intro}

\indent In recent years, the exponential advancements of AI have been largely driven by the massive amounts of data. In computer vision, with the emergency of SA-1B~\cite{Kirillov_2023_ICCV} and LAION-5B~\cite{schuhmann2022laion}, models like SAM~\cite{Kirillov_2023_ICCV} and Stable Diffusion~\cite{rombach2022high} have greatly benefited from these large volumes of data, enabling zero-shot transfer to downstream tasks.  Subsequently, Objaverse~\cite{deitke2023objaverse, deitke2023objaversexl} and MVImgNet~\cite{yu2023mvimgnet} break barriers of 3D data collection with large-scale synthetic 3D assets and real-world multi-view capture, which support Zero123\cite{liu2023zero} and LRM~\cite{hong2023lrm} models to achieve impressive generalization ability of multi-view or 3D reconstruction. However, comparable progress on human-centric tasks still remained elusive due to the limited scale of 3D human data.

Compared to collecting 3D object datasets, capturing high-quality and large-scale 3D human avatars is more time-consuming in the same order of scale. Existing 3D human datasets can be categorized into two distinct representations: 3D human scans and multi-view human images. While 3D human scan data~\cite{RenderPeople, yu2021function4d} provides accurate geometric shapes, it comes with high acquisition costs which leads to limited data scale. Conversely, multi-view capture provides an easier way to collect 3D human data. Previous multi-view human datasets~\cite{peng2021neural,ionescu2013human3,li2021ai} involve only a few dozen human subjects. Recent advances in multi-view human performance data~\cite{cheng2022generalizable,cheng2023dna} narrow the gap of data scarcity which provides more diverse and representative human data for establishing reasonable benchmarks. To ensure comprehensiveness, it is necessary for these datasets to consider the complex clothing and the uncommon human-object interaction. However, incorporating these factors introduces complexities for scaling up the dataset.


To scale up the 3D human data, we present \textbf{MVHumanNet}, a large-scale multi-view human performance capture dataset. Our dataset primarily focuses on casual clothing commonly found in everyday life, enabling 
to easily expand the scale of human data collection. For the hardware setup, we establish two 360-degree indoor systems equipped with 48 and 24 calibrated RGB cameras, respectively, to capture high-fidelity videos with resolutions up to 12MP (4096 × 3000)  and 5MP (2048 × 2448). Considering the capture of human data, we intend to cover a wide range of attributes among human subjects, including age, body shape, motion, as well as the colors, types, and materials of dressing, enabling our dataset as diverse as possible. Furthermore, we also design 500 motion types to guarantee coverage of daily scenarios. Overall, we invite \textbf{4,500} individuals to participate in the data capture process. Each participant is recorded in two distinctive outfits (\textbf{9,000} in total) and at least seven different motion sequences. Thanks to the targeted collection of everyday clothing,  data capture for each participant has been accomplished efficiently within six months. Eventually, the full dataset comprises an extensive collection of \textbf{60,000} motion sequences with over \textbf{645 million} frames. Compared with the existing multi-view human datasets~\cite{ionescu2013human3, peng2021neural, cheng2022generalizable, isik2023humanrf}, MVHumanNet provides a significantly larger number of human subjects and outfits than previously available. Furthermore, MVHumanNet surpasses the recently proposed DNA-Rendering~\cite{cheng2023dna} dataset by an order of magnitude in terms of motion and frame data. The detailed comparisons between MVHumanNet and other relevant datasets are shown in  ~\cref{tab:dataset_comparison}.

In order to benefit downstream human-centric tasks, we also provide essential annotations including action labels, camera intrinsics and extrinsics,  human masks, 2D/3D keypoints,  SMPL/SMPLX parameters and text descriptions to enhance the applicability of our dataset.  To thoroughly explore the capabilities of our dataset, we carefully design four pilot experiments: \textbf{a)} view-consistent action recognition, \textbf{b)} NeRF reconstruction for human, \textbf{c)} text-driven view-unconstrained human image generation, and \textbf{d)} 2D view-unconstrained human image and 3D avatar generation. First, leveraging the multi-view nature of human capture data, we can achieve more accurate view-consistent action recognition and enhance the generalization capability of NeRF as the data scale increases. Furthermore, the unprecedented scale of subjects and outfits, along with pose sequences and paired textual descriptions, allows us to finetune a remarkable text-driven, pose-conditioned high-quality human image generation model. Finally, through the exploitation of multi-view human images on a large scale, we can obtain 2D/3D full-body human generative models with promising results.
The aforementioned experiments reveal the promise and opportunities with the large-scale MVHumanNet datasets to boost a wide range of digital human applications and inspire future research.


In summary, the main contributions of our work include: 
\begin{itemize} [itemsep=2pt,topsep=2pt,parsep=0pt, leftmargin=2em]
    \item We present the largest multi-view human capture dataset, which is nearly ten times larger than the recently proposed DNA-Rendering dataset in terms of human subjects, motion sequences, and frames.
    \item We conduct several pilot studies that demonstrate the proposed MVHumanNet can support various downstream human-centric tasks for effective applications.
    \item We believe that MVHumanNet opens up new possibilities for research in the field of 3D digital human. 
\end{itemize}


\begin{table*}[htbp]
    \centering
    \resizebox{1.0\textwidth}{!}{%
    \begin{tabular}{c|ccc|ccccc|c}
        \toprule
        Dataset & Age & Cloth & Motion & \#ID & \#Outfit & \#Actions & \#View & \#Frames & Resolution\\
        \midrule
        Human3.6M~\cite{ionescu2013human3} & \textcolor{darkred}{\XSolidBrush} & \textcolor{darkred}{\XSolidBrush} & \textcolor{darkgreen}{\textcolor{darkgreen}{\CheckmarkBold}} & 11  &  11 & 17 & 4 & 3.6M & 1000P \\
        CMU Panoptic~\cite{joo2015panoptic} & \textcolor{darkgreen}{\CheckmarkBold} & \textcolor{darkred}{\XSolidBrush} & \textcolor{darkgreen}{\CheckmarkBold} & 97  & 97 & 65 & 31 & 15.3M & 1080P \\
       MPI-INF-3DHP~\cite{mehta2017monocular} & \textcolor{darkred}{\XSolidBrush} & \textcolor{darkred}{\XSolidBrush} & \textcolor{darkgreen}{\CheckmarkBold} & 8  & 8 & $-$  & 14 &  1.3M & \cellcolor{lightgray}2048P \\
        NHR~\cite{wu2020multi} & \textcolor{darkred}{\XSolidBrush} & \textcolor{darkred}{\XSolidBrush} & \textcolor{darkgreen}{\CheckmarkBold} & 3  & 3 & 5 & 80 & 100K & \cellcolor{lightgray}2048P \\
        ZJU-MoCap~\cite{peng2021neural} & \textcolor{darkred}{\XSolidBrush} & \textcolor{darkred}{\XSolidBrush} & \textcolor{darkgreen}{\CheckmarkBold} & 10  & 10 & 10 & 24 & 180K & 1024P \\
        Neural Actor~\cite{liu2021neural} & \textcolor{darkred}{\XSolidBrush} & \textcolor{darkred}{\XSolidBrush} & \textcolor{darkgreen}{\CheckmarkBold} & 8  & 8 & $-$  & 11$\sim$100 & 250K & 1285P \\
        HUMBI~\cite{yu2020humbi} & \textcolor{darkgreen}{\CheckmarkBold} & \textcolor{darkgreen}{\CheckmarkBold} & \textcolor{darkred}{\XSolidBrush} & 772  & 772  & $-$ & \cellcolor{lightgray}107 & 26M & 1080P \\
        AIST++~\cite{li2021ai} & \textcolor{darkred}{\XSolidBrush} & \textcolor{darkred}{\XSolidBrush} & \textcolor{darkred}{\XSolidBrush} & 30  & 30 & $-$ & 9 & 10.1M & 1080P \\
        THuman 4.0~\cite{zheng2022structured} & \textcolor{darkred}{\XSolidBrush} & \textcolor{darkred}{\XSolidBrush} & \textcolor{darkgreen}{\CheckmarkBold} & 3  & 3 & $-$ & 24 & 10K & 1150P \\
        HuMMan~\cite{cai2022humman} & \textcolor{darkred}{\XSolidBrush} & \textcolor{darkgreen}{\CheckmarkBold} & \textcolor{darkgreen}{\CheckmarkBold} & \cellcolor{lightgray}1000  & 1000 & \cellcolor{lightgray}500 & 10 & 60M & 1080P \\
        GeneBody~\cite{cheng2022generalizable} & \textcolor{darkgreen}{\CheckmarkBold} & \textcolor{darkgreen}{\CheckmarkBold} & \textcolor{darkgreen}{\CheckmarkBold} & 50  & 100 & 61 & 48 & 2.95M & \cellcolor{lightgray}2048P \\
        ActorsHQ~\cite{isik2023humanrf} & \textcolor{darkred}{\XSolidBrush} & \textcolor{darkred}{\XSolidBrush} & \textcolor{darkgreen}{\CheckmarkBold} & 8  & 8 & 52 & \cellcolor{deeppeach}160 & 40K & \cellcolor{deeppeach}4096P \\
        DNA-Rendering~\cite{cheng2023dna} & \textcolor{darkgreen}{\CheckmarkBold} & \textcolor{darkgreen}{\CheckmarkBold} & \textcolor{darkgreen}{\CheckmarkBold} & \cellcolor{lightgray}500  & \cellcolor{lightgray}1500 & \cellcolor{deeppeach}1187 & 60 & \cellcolor{lightgray}67.5M & \cellcolor{deeppeach}4096P \\
        \midrule
        \textit{\textbf{MVHumanNet(Ours)}} & \textcolor{darkgreen}{\CheckmarkBold} & \textcolor{darkgreen}{\CheckmarkBold} & \textcolor{darkgreen}{\CheckmarkBold} & \cellcolor{deeppeach}4500  & \cellcolor{deeppeach}9000 & \cellcolor{lightgray}500 & 48 & \cellcolor{deeppeach}645.1M & \cellcolor{deeppeach}4096P \\
        \bottomrule
    \end{tabular}%
    }
        \captionsetup{skip=6pt} 
     \caption{\textbf{Dataset comparison on existing multi-view human-centric datasets.} MVHumanNet provides a significantly larger number of human subjects and outfits than previous datasets available, regarding the number of identities (\#ID), outfits in total (\#Outfit) and frames of images (\#Frames). Attributes among humans, including age, cloth and motion are covered (denoted by \textcolor{darkgreen}{\text{\ding{51}}} for inclusion and \textcolor{darkred}{\text{\ding{55}}} for exclusion.). Cells highlighted in {\begin{tikzpicture} [scale=1.0] \fill[deeppeach] (0,0) rectangle (1em,1em); \end{tikzpicture}}   {\begin{tikzpicture}[scale=1.0]  \fill[color=lightgray] (0ex,0ex) rectangle (1em,1em);  \end{tikzpicture} denotes the dataset with best and second-best feature in each column.}
     \vspace{-3mm}
     }
    \label{tab:dataset_comparison}
\end{table*}

\section{Related Work}
\noindent \textbf{3D Human Reconstruction and Generation.} 
Recently, we have witnessed impressive performance in the field of image generation, 3D reconstruction and novel view synthesis in computer vision community with the emergency of Generative Adversarial Networks (GANs)~\cite{goodfellow2014gan, isola2017image, karras2019style}, Neural Implicit Function~\cite{park2019deepsdf, mescheder2019occupancy, chen2019learning} and Neural Radiance Field (NeRF)~\cite{mildenhall2020nerf, muller2022instant}. These successes inspire subsequent works~\cite{fruhstuck2022insetgan, saito2019pifu, peng2021neural, kwon2021neural}  to extend reconstruction and generation tasks to high-fidelity clothed full-body humans.
Many efforts have also been made to combine 2D GANs with NeRF representations for 3D-aware, photo-realistic image synthesis. EG3D~\cite{chan2022efficient} proposes the 3D-aware generation of multi-view face images by introducing an efficient tri-plane representation for volumetric rendering.   GET3D~\cite{gao2022get3d} utilizes two separate latent codes to generate the SDF and texture field, enabling the generation of textured 3D meshes. EVA3D~\cite{hong2022eva3d} extends EG3D to learn generative models with human body priors for 3D full-body human generation from a collection of 2D images. HumanGen~\cite{jiang2023humangen} and Get3DHuman~\cite{xiong2023Get3DHuman} further incorporate the priors of StyleGAN-Human~\cite{fu2022styleganhuman} and PIFuHD~\cite{saito2020pifuhd} for generative human model construction. In addition, Text2Human~\cite{jiang2022text2human} and AvatarClip~\cite{hong2022avatarclip} explore to leverage the powerful vision-language model CLIP~\cite{radford2021learning} for text-driven 2D and 3D human generation.  However, the reconstruction, generation and novel view synthesis works can only utilize limited real-world human data, which consequently affects the generalizability of their models. Moreover, the current methods of human generation often train their models on datasets comprising only front-view 2D human images~\cite{liu2016deepfashion, fu2022styleganhuman} or monocular human videos~\cite{zablotskaia2019dwnet}. Unfortunately, these approaches fail to produce satisfactory results when altering the input image across various camera viewpoints. In this work, we provide the current largest scale of  multi-view human capture images along with text descriptions to facilitate 3D human-centric tasks.

\noindent \textbf{3D Human Scanning Datasets.} Understanding human actions and reconstructing detailed body geometries with realistic appearances are challenging tasks that require high-quality and large-scale human data. Early works~\cite{bogo2014faust, bogo2017dynamic, zhang2017detailed} in this field provide dynamic human scans but with limited data consisting of only a few subjects or simple postures. Parallel works such as Northwestern-UCLA~\cite{wang2014cross} and NTU RGB+D series~\cite{liu2019ntu, shahroudy2016ntu} utilize more affordable Kinect sensors to obtain depth and human skeleton data, enabling the capture of both appearance and action cues. However, due to the limitations in the accuracy of Kinect sensors, these datasets are inadequate for precise human body modeling. Subsequently, AMASS~\cite{AMASS:ICCV:2019} further integrates traditional motion capture datasets~\cite{CMU_mocap, SFU_mocap} and expands them with fully rigged 3D meshes to facilitate advancements in human action analysis and body modeling research. With the emergency of learning-based digital human techniques, relevant algorithms~\cite{saito2019pifu, xiu2022icon, saito2020pifuhd, chen2021snarf} heavily rely on human scan datasets with high-fidelity 3D geometry and corresponding images. Several studies~\cite{zheng2019deephuman, yu2021function4d, zheng2021deepmulticap, shen2023x, han2023high, ma2020learning} capture their own datasets and release the data to the public for research purposes. Additionally, there are several commercial scan datasets~\cite{RenderPeople, Twindom, AXYZ, 3dpeople} that are well-polished and used for research to ensure professional quality. These datasets play a foundational role in bridging the gap between synthetic virtual avatars and real humans. However, the aforementioned datasets typically exhibit a bias towards standing poses due to the complicated capture procedure and cannot afford for large-scale data collection. 

\noindent \textbf{Multi-view Human Capturing Datasets.} Multi-view capture holds an indispensable role in computer vision, serving as a fundamental technique for AR/VR and 3D content production. Prior works~\cite{vlasic2008articulate, vlasic2009dynamic} present multi-view stereo systems to collect multi-view human images and apply multi-view constraints to reconstruct 3D virtual characters.  Human3.6M~\cite{ionescu2013human3} captures numerous 3D human poses using a marker-based motion capture system from 4 cameras. MPI-INF-3DHP~\cite{mehta2017monocular} annotates both 3D and 2D pose labels for human motion capture in a multi-camera studio.  CMU Panoptic~\cite{joo2015panoptic} presents a massively multiview system consisting of 31 HD Cameras to capture social interaction and provides 3D keypoints annotations of multiple people. HUMBI~\cite{yu2020humbi} collects local human behaviors such as gestures, facial expressions, and gaze movements from multiple cameras. AIST++~\cite{li2021ai,tsuchida2019aist}  is a dance database that contains various 3D dance motions reconstructed from real dancers with multi-view videos. These datasets primarily focus on human activity motions ranging from daily activities to professional performances, rather than factors related to identity,  cloth texture and body shape diversity. With the recent progress of neural rendering techniques, NHR~\cite{wu2020multi}, ZJU-Mocap~\cite{peng2021neural}, Neural Actor~\cite{liu2021neural, habermann2021real, habermann2020deepcap} and THuman4.0~\cite{zheng2022structured} present their multi-view human dataset for evaluating the proposed human rendering algorithms. HuMMan~\cite{cai2022humman} and Genebody~\cite{cheng2022generalizable} expand the diversity of pose actions and body shapes for human action recognition and modeling. ActorsHQ~\cite{isik2023humanrf} uses dense multi-view capturing for photo-realistic novel view synthesis but is limited to 16 motion sequences and 8 actors.  
Recently, with the presence of the large-scale synthetic data and real captures from Objaverse~\cite{deitke2023objaverse, deitke2023objaversexl} and MVImgNet~\cite{ yu2023mvimgnet}, several methods~\cite{liu2023zero, hong2023lrm} have made remarkable strides in the direction of open-world 3D reconstruction and generation. The concurrent work, DNA-Rendering~\cite{cheng2023dna} emphasizes the comprehensive benchmark functionality, but it encounters challenges in expanding the dataset to a larger scale due to the consideration of unusual human-object interactivity and clothes texture complexity. Differing from these efforts, we take a significant step forward in scaling up the human subjects and outfits, leading to the creation of MVHumanNet, the multi-view human capture dataset on the largest scale.

\section{MVHumanNet}
In this section, we provide a comprehensive overview of the core features of \thename, with a focus on dataset construction. 
We discuss the hardware capture system, data collection arrangements, dataset statistics, and data pre-processing. 
\cref{data_sys} provides an illustration to the fundamental aspects of the data acquisition system. 
This part specifically outlines the key components of the hardware capture system and its capabilities. \cref{data_cap} delves into the actual data acquisition process, providing detailed information on personnel arrangement and the protocols followed during data collection. 
This section elucidates the steps taken to ensure the accuracy and consistency of the acquired data.
Finally, in~\cref{data_ann}, we present a comprehensive framework that combines manual annotation and existing algorithms to obtain diverse and rich annotations for \thename. This framework enhances the applicability of our dataset for various research purposes.

\subsection{Multiview Synchronized Capture System}\label{data_sys}
We collected all the data using two sets of synchronized indoor video capture systems. In this section, we provide a detailed account of one system, while supplementary materials contain additional information about the second system. The primary framework of the capture system consists of 48 high-definition industrial cameras with a resolution of 12MP. These cameras are arranged in a multi-layer structure resembling a 16-sided prism, as shown in~\cref{fig:teaser}. 
The collection system has approximate dimensions of 2.4 meters in height and a diameter of roughly 4.5 meters.
Each prism within the system is equipped with three 4K high-definition industrial cameras positioned at different heights.
The lenses of each camera are meticulously aligned towards the center of the prism. 
To ensure clear image capture from different perspectives, we have placed light sources at the center of each edge of the system.
During the data collection process, the frame rate of all cameras is set to 25 frames per second, enabling the capture of smooth and detailed motion sequences. 
For more comprehensive technical details, please refer to the supplementary materials.

\subsection{Data Capture and Statistics}\label{data_cap}

\noindent\textbf{Data Capture} 
To capture the wide range of dressing habits observed in people's daily lives, we establish a comprehensive process for performer recruitment and data collection.
Specifically, at regular intervals, we release targeted recruitment requests to the public based on the statistics derived from the already collected clothing data. 
This strategy aims to enhance the diversity of clothing styles and colors for more reasonable human data distributions to achieve more reasonable human data distributions.
In accordance with the clothing requirements, each performer is instructed to bring two sets of clothing to the capture system.  Prior to the beginning of the capturing, performers randomly select 12 sets of actions from a predefined pool of 500 actions. 
Subsequently, they enter the capture system and sequentially perform the first six sets of actions, following instructions provided by the collection personnel. Each action is performed at least once on both the left and right sides for complete execution of the human performance capture.
Upon completing the sixth set of actions, the performer finishes the first collection session by extending their hands to an A-pose and rotating in place twice. Subsequently, the performer changes outfit and repeats the same process to complete the remaining six sets of actions with rotations in place.

\begin{figure}[t]
\begin{center}
\includegraphics[width=0.99\linewidth]{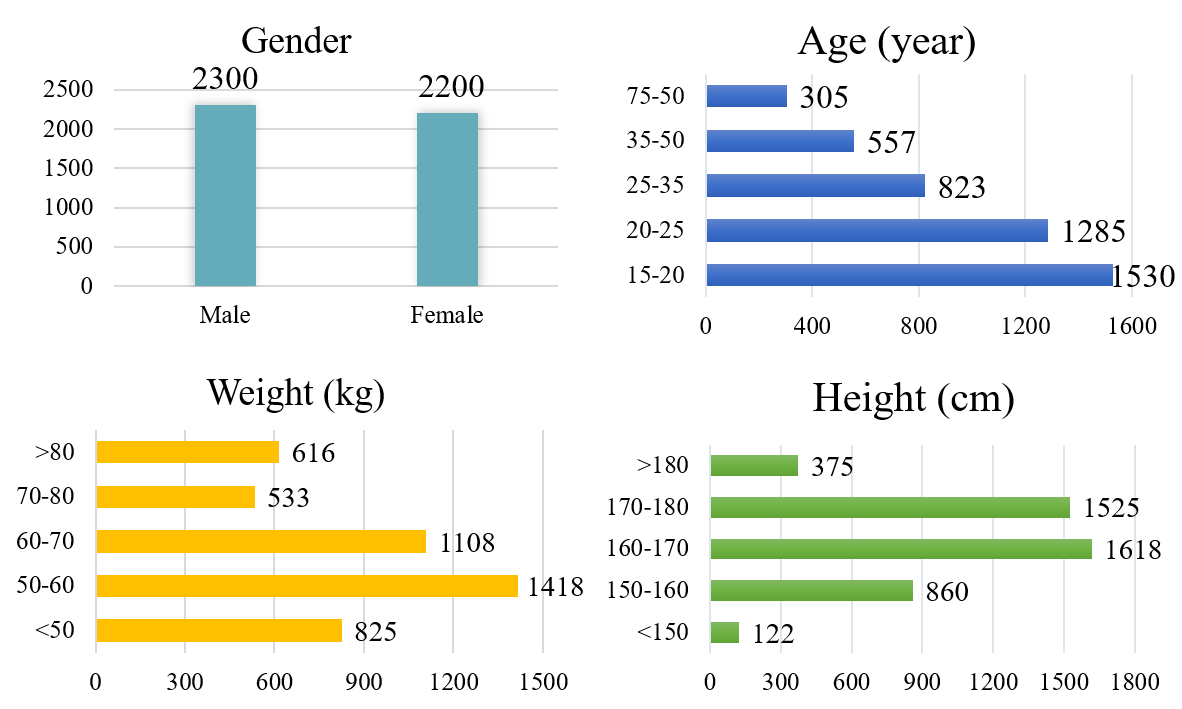}
\end{center}
\vspace{-6mm}   
\caption{
\textbf{The distribution of performers' attributes.} The gender, age, weight, and height of performers are recorded and carefully controlled. The statistical analysis of these attributes reflects a diverse range among the performers involved in \thename.}
\label{fig:0_human}
\vspace{-5mm}
\end{figure}

\begin{figure}
\begin{center}
\includegraphics[width=0.99\linewidth]{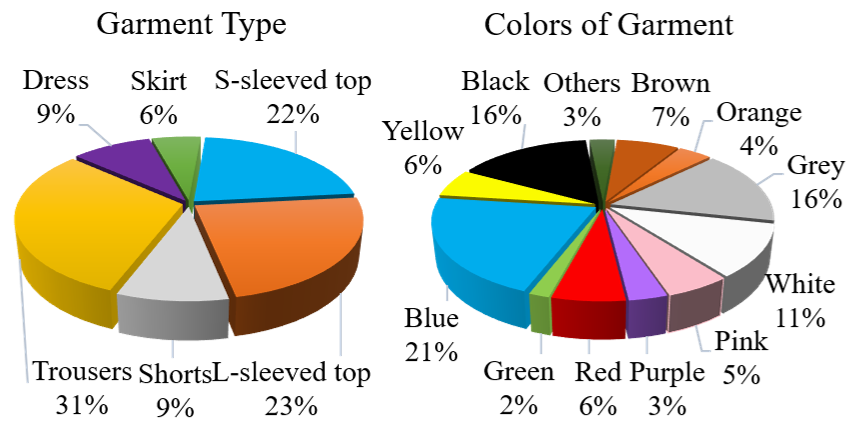}
\end{center}
\vspace{-6mm} 
\caption{
\textbf{The garment type and color distribution of outfits of performers.} Diverse colors and types of dressing are required for each invited performer. The statistical results show the wide coverage of daily clothes.
}
\vspace{-5mm}
\label{fig:0_garment}
\end{figure}

\noindent\textbf{Data Statistics}
The essential statistics of our dataset are shown in~\cref{fig:0_human} and~\cref{fig:0_garment}.
\thename~comprises a total of 4,500 unique identities with a equitable distribution of 2,300 male and 2,200 female individuals, ensuring a balanced representation of genders.
Participants are required to fall within the age range of 15 to 75 years old. 
This age range is chosen to encompass a wide spectrum of performers while considering the potential impact of age on the quality and capabilities of their actions.
Conversely, no restrictions are imposed on performers' weight or height, as these variables are deemed to have minimal impact on the data collection process. 
By not imposing such limitations, we aim to capture a more diverse and realistic representation of subjects in the dataset, allowing for a broader range of body types and proportions.
Our dataset boasts the largest number of unique identities and garment items when compared to existing multi-view human dataset . 
It encompasses a wide range of everyday clothing styles and colors that are commonly available in real-world scenarios.



\subsection{Data Annotation}\label{data_ann}
\begin{figure}[tb]
\begin{center}
\includegraphics[width=1.0\linewidth]{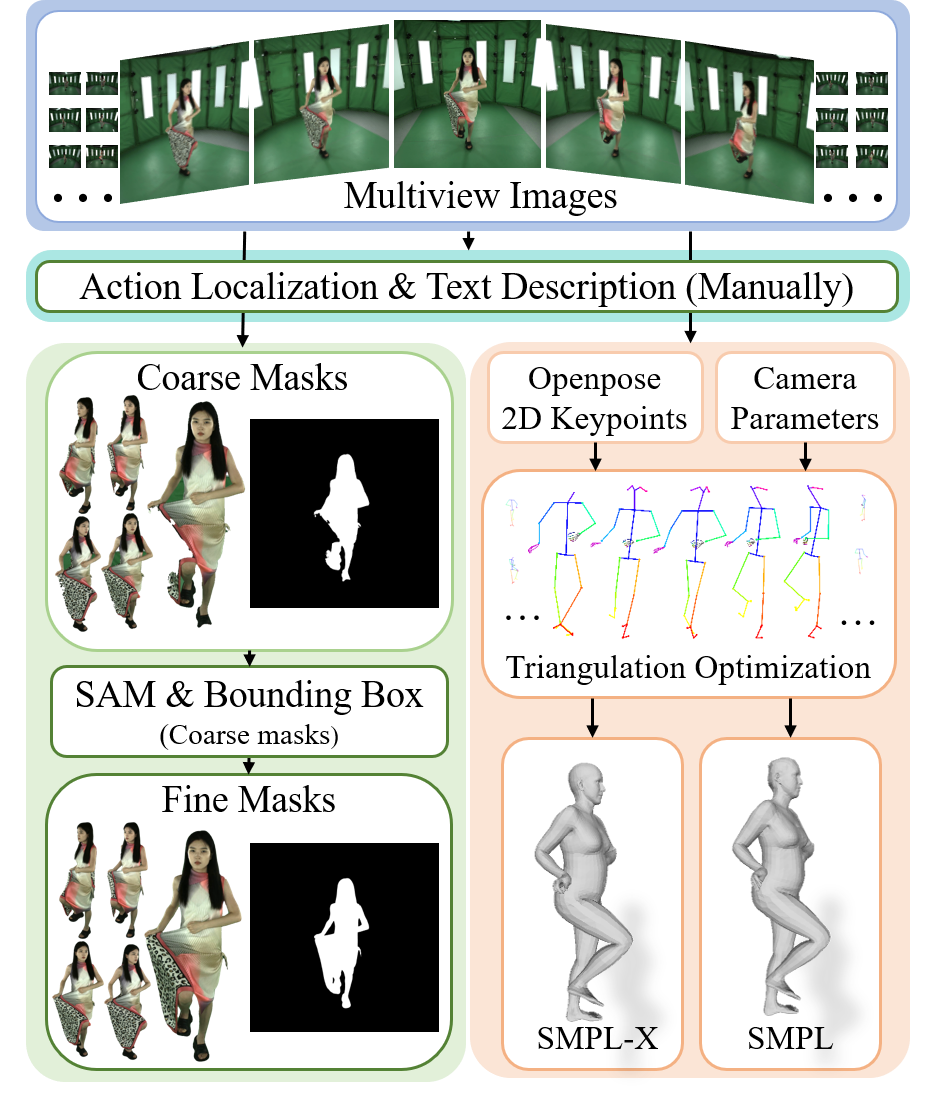}
\end{center}
\vspace{-8mm}
\caption{
\textbf{Data annotation pipeline.} The manual and automatic annotation pipeline for action localization, text description, masks, 2D/3D keypoints and parametric models.}
\label{fig:process}
\vspace{-5mm}
\end{figure}

To enable the advancement of applications in 2D/3D human understanding, reconstruction and generation, our dataset offers comprehensive and diverse annotations alongside the raw data.
These annotations include action localization, attribute description, human masks, camera calibrations, 2D/3D skeleton, and parametric model fitting. 
The annotation pipeline, as depicted in~\cref{fig:process}, provides an overview of the entire process.

\noindent\textbf{Manually Annotation} 
Before capturing human data, we collect the cloth color and dress type of each performer in the registration table for further manual textual description. During the data collection process, we ensure a continuous flow as performers execute a sequence of six distinct actions along with in-place rotations. 
Subsequently, after the recording session, we manually mark the breakpoints for each action, accurately documenting the start and end  of each action sequence.  
Moreover, the supplementary materials provide comprehensive records and annotations of the performers' basic attributes and outfits. 
For further details, please refer to the supplementary materials.

\noindent\textbf{Human Mask Segmentation}
\thename~comprises approximately 645 million images of individuals captured from various perspectives. 
Manual segmentation of such a massive image collection is obviously infeasible. 
To tackle this challenge, we propose a hierarchical automated image segmentation approach based on off-the-shelf segmentation algorithms. 
Our approach follows a coarse-to-fine segmentation strategy. 
Initially, we employ the RVM~\cite{rvm} to obtain efficient rough segmentation results. Subsequently, the rough segmentation outcomes are utilized to generate a bounding box of the performer, which serves as a prompt for the SAM~\cite{Kirillov_SAM}  to produce higher-quality masks. 
In~\cref{fig:process}, the bottom-left region presents a comparison between the coarse and fine segmentation results. 
Notably, the masks generated by SAM exhibit significant superiority to those generated by RVM. 
Nevertheless, RVM remains crucial as it provides a rough bounding box, ensuring that the fine stage SAM segmentation primarily focuses on the individual rather than other elements.  

\noindent\textbf{Camera Calibration}
We utilized a commercial solution based on CharuCo boards to achieve fast and efficient camera calibration. 
Specifically, we position a CharuCo patterned calibration board at the central location of the capture studio.
This ensures that each camera can capture a clear and complete view of the calibration board. With the aid of specific software, we obtain the intrinsic, extrinsic parameters and distortion coefficient for each camera. 
We also carefully adjust other parameters, such as lighting, exposure, and white balance to capture high-quality data.

\noindent\textbf{2D/3D Skeleton and Parametric Models}
Following the previous works~\cite{cai2022humman, cheng2023dna, cheng2022generalizable} and with the goal of facilitating extensive research and applications in 3D digital human community, we conducted pre-processing on the entire dataset to obtain corresponding 2D/3D skeletons and two parameterized models. 
The processing pipeline is visually depicted in the bottom-right section of~\cref{fig:process}. 
Specifically, we employed the OpenPose~\cite{cao2017openpose} to predict 2D skeletons for each frame of the images.
Leveraging the calibrated camera parameters, multi-view 2D skeletons, and optimization algorithms~\cite{easymocap}, we derived the 3D skeletons from multi-view triangulation. Finally, SMPL/SMPLX parameters are fitted with the constrains of multi-view 2D keypoints and 3D skeletons.
All these labeled data support MVHumanNet to be applied to various tasks.

\section{Experiments}
In this section, we present a comprehensive series of exploratory experiments conducted in the human action understanding, reconstruction, and generation tasks. Specifically, \cref{action_task} focuses on showcasing experiments pertaining to view-consistent action recognition. 
As the dataset expands from single-view 2D data to multi-view 3D data, existing algorithms may encounter new challenges. 
In~\cref{nerf_task}, we demonstrate experiments on generalizable NeRF (Neural Radiance Fields) reconstruction approaches, emphasizing the augmented model performance and generalization capabilities resulting from the increased availability of data. 
At last, in~\cref{text_task} and~\cref{avatar_task}, we delve into recent research tasks, specifically text-driven view-unconstrained image generation and 3D human avatar generative model.
Taking into account the size of the dataset, hardware limitations, and data annotation constraints, we performed experiments utilizing 62{\%} of the available data. More precisely, we employed 2800 identities, each representing a unique set of attire, amounting to a total of 5500 sets. Within this subset, 10{\%} of the data was reserved exclusively for testing purposes.

\begin{table}[b]
\centering
\small
\vspace{-2mm}
\def\arraystretch{1} \tabcolsep=0.4em 
\begin{tabular}{r|r|c|c|c}
    \toprule
    & \makecell{Train\\ views}& CTR-GCN\cite{chen2021CRTGCN} & InfoGCN\cite{chi2022infogcn} & FR-Head\cite{zhou2023FRHEAD}  \\
    \midrule
    \makecell{Top-1\\(\%)$\uparrow$ }& 
    \makecell{1-view\\ 2-views\\  4-views\\ 8-views} &  
    \makecell{33.85\\60.33\\72.16\\ 76.73 }  & 
    \makecell{25.23\\55.89\\73.59\\ 76.55}  &  
    \makecell{30.25\\59.16\\71.74\\ 78.19}  \\
    \midrule
    \makecell{Top-5\\(\%)$\uparrow$ }& \makecell{1-view\\ 2-views\\ 4-views\\ 8-views} & 
    \makecell{51.08\\80.09\\88.32\\ 91.34}  & 
    \makecell{37.14\\75.00\\89.02\\ 91.00}  &  
    \makecell{50.59\\78.80\\88.67\\ 92.45}  \\
    \bottomrule
\end{tabular}
\vspace{-3mm}
\caption{Performance comparison of skeleton-based action recognition SOTA methods on \thename. With the increase of the views, the accuracy of the action prediction increases together.}
    \label{tab:action_experiments}
\end{table}

\subsection{View-consistent Action Recognition }\label{action_task}
\thename~provides action labels with 2D/3D skeleton annotations, which can verify its usefulness on action recognition tasks. 
To simulate real-world scenarios, we employed single-view 2D skeletons as input and conducted tests on a multi-view test set that accurately represented real scenes. 
Our experimentation involved 8 viewpoints spaced at 45-degree intervals. 
The training data encompassed approximately 4000 outfits, while the testing data included 400 outfits, covering a total of 500 action labels. 
The results, presented in~\cref{tab:action_experiments}, reveal that the accuracy of action estimation was notably low for a single viewpoint, achieving a top-1 accuracy of only around 30\%. 
However, as the number of input viewpoints increased, the accuracy of action estimation exhibited a significant improvement, peaking at 78.19\%. 
Given that the dataset covers a comprehensive range of daily full-body actions, we possess confidence in its efficacy for facilitating diverse understanding tasks.
Considering the challenges associated with acquiring 3D skeletons in everyday life, see supplementary for the results of 3D skeleton-based action recognition.

\subsection{NeRF Reconstruction for Human}\label{nerf_task}

\begin{table}[tb]
    \resizebox{\columnwidth}{!}{
    \begin{tabular}{c|ccc|ccc}
        \toprule
        {\multirow{2}*{\begin{tabular}[c]{@{}c@{}}Number of \\ outfits\end{tabular}}}   &  \multicolumn{3}{c|}{IBRNet~\cite{wang2021ibrnet} }   & \multicolumn{3}{c}{GPNeRF~\cite{chen2022geometry}}   \\
         &  PSNR $\uparrow$ &  SSIM $\uparrow$ &  LPIPS $\downarrow$ &  PSNR  $\uparrow$ &  SSIM $\uparrow$ &  LPIPS $\downarrow$ \\
        \midrule
             100     & 26.05  & 0.9571  & 0.0555  &  23.27 & 0.8688 &  0.2077 \\
            2000    & 27.45  & 0.9638  & 0.0486  & 24.14 & 0.8779 & 0.2137 \\
            5000    & 29.00 & 0.9706 & 0.0377  & 24.69 & 0.8878 & 0.1961 \\
        \bottomrule
    \end{tabular}
    }
    \vspace{-3mm}
    \caption{\small \textbf{Quantitative comparison of generalizable NeRFs with different scales of data for training.} We compare the results of methods with human prior and without human prior. We refer human prior to the commonly used SMPL model.}
     \vspace{-4mm}
    \label{tab:gene_human_nerf}
\end{table}

\begin{figure}[tb]
\begin{center}
\includegraphics[width=0.99\linewidth]{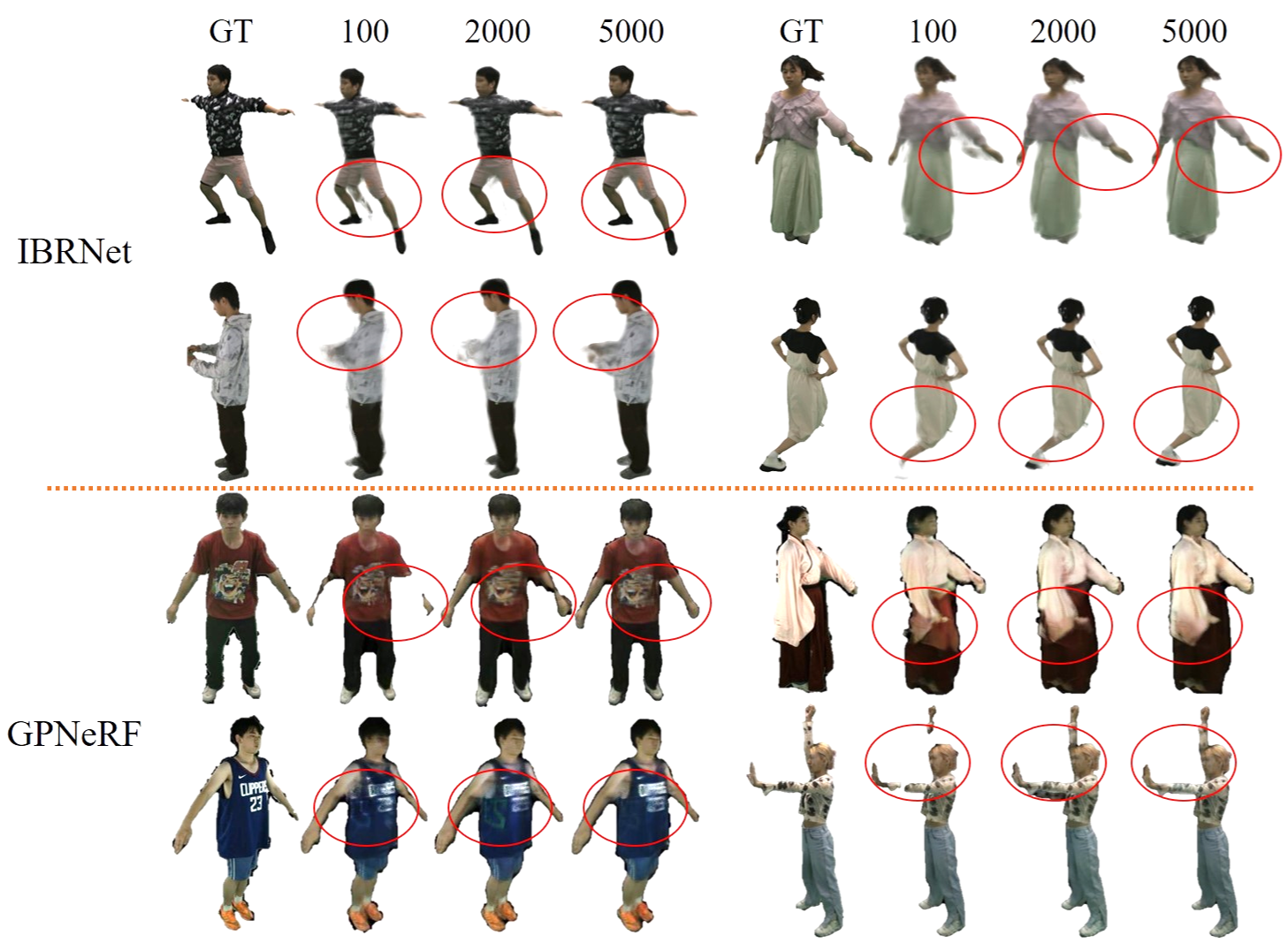}
\end{center}
\vspace{-5mm}
\caption{\textbf{The novel view synthesis results of IBRNet and GPNeRF on unseen data of MVHumanNet.} \textbf{GT} means ground truth. The number of \textbf{100}, \textbf{2000}, and \textbf{5000} indicate the respective quantities of outfits utilized during the training process.}
\vspace{-4mm}
\label{fig:nerf_vis}
\end{figure}

MVHumanNet can also be applied to NeRF reconstruction for human. Currently, human-centric methods, \eg GPNeRF \cite{chen2022geometry}, are developed in the context of lacking multi-view human data and their performance is still far from satisfactory on more diverse testing cases. We hope our proposed MVHumanNet can motivate more extensive studies of generalizable NeRF for human with sufficiently large scale of data.  We empirically explore the performance of two distinct generalizable NeRFs methods,  IBRNet \cite{wang2021ibrnet} which is designed for general scenes and GPNeRF \cite{chen2022geometry} which relies on human prior (\ie SMPL~\cite{loper2023smpl}), using varying amounts of data for training. In our experiment, both approaches utilize four evenly distributed views as input and inference the novel view results. The quantitative comparisons of the outcomes are presented in Tab.~\ref{tab:gene_human_nerf}, while the visualization results can be found in Fig.~\ref{fig:nerf_vis}. Experimental results confirm that as the training data increases, the model exhibits enhanced generalization capabilities for new cases, especially when facing rare poses and complex garments. Moreover, we provided empirical evidence that MVHumanNet can also serve for pretraining strong models, facilitating methods to perform better on out-of-domain scenarios. The corresponding results are presented in Tab.~\ref{tab:cross_dataset_nerf}. Please note that the quantitative results of IBRNet \cite{wang2021ibrnet} and GPNeRF \cite{chen2022geometry} cannot be directly compared, as they have different evaluation settings. More detailed explanations are in Supp.

\begin{table}[tb]
    \resizebox{\columnwidth}{!}{
    \begin{tabular}{c|ccc|ccc}
        \toprule
          {\multirow{2}*{Method}}  &  \multicolumn{3}{c|}{IBRNet~\cite{wang2021ibrnet}} & \multicolumn{3}{c}{GPNeRF~\cite{chen2022geometry}} \\
          &  PSNR $\uparrow$ &  SSIM $\uparrow$ & LPIPS $\downarrow$ &  PSNR $\uparrow$  &  SSIM $\uparrow$ & LPIPS $\downarrow$ \\
        \midrule
          Train from scratch & 28.06 &  0.9679 & 0.0437        & 20.95 & 0.9049 & 0.1809\\
          w/o fintune & 27.48 & 0.9663 & 0.0440 & 20.15 & 0.8921 & 0.2050 \\
          w/ fintune  & 29.46 & 0.9734 & 0.0323 & 21.89 & 0.9252 & 0.1364  \\
        \bottomrule
    \end{tabular}
    }
    \vspace{-3mm}
    \caption{\small \textbf{Using MVHumanNet to pretrain a strong model}. We first train the representative methods on MVHumanNet, and then finetune the trained models on the train set of HuMMan~\cite{cai2022humman}. We compare the performance of the finetuned models and models trained from scratch on the test set of HuMMan.}
    \vspace{-4mm}
        \label{tab:cross_dataset_nerf}
\end{table}

\begin{figure}[tb]
\begin{center}
\includegraphics[width=0.99\linewidth]{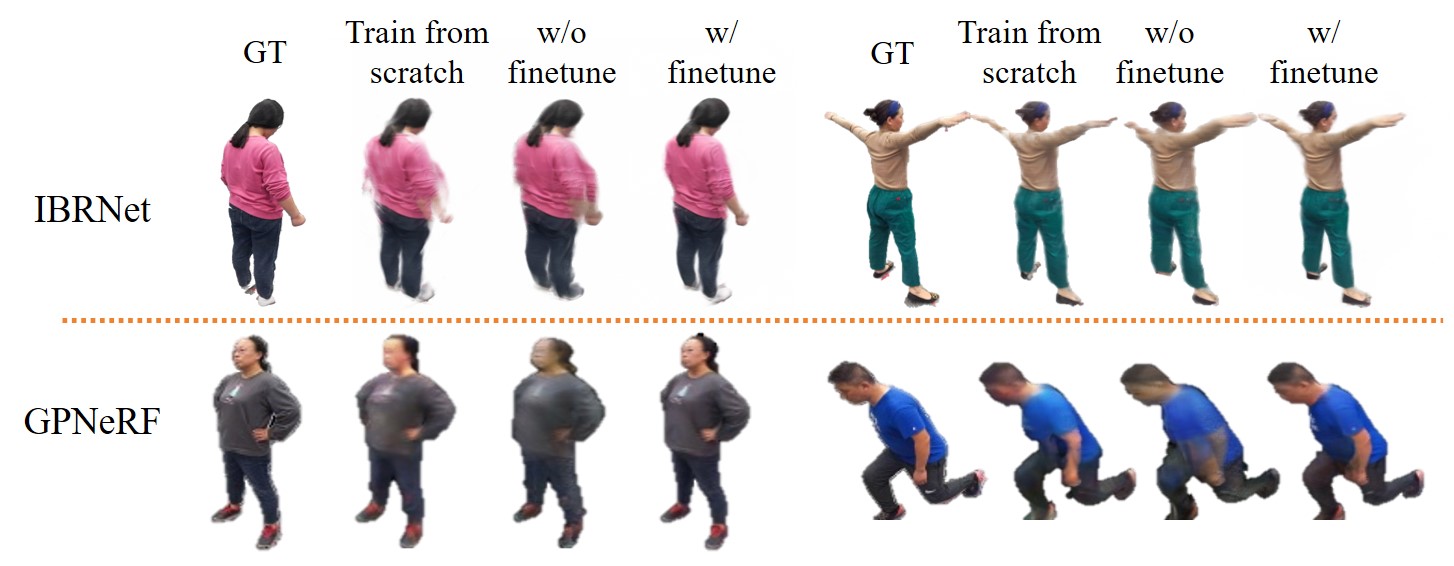}
\end{center}
\vspace{-7mm}
\caption{\textbf{Qualitative comparison of IBRNet and GPNeRF on the test set of HuMMan.} Without finetuning, the models only trained on MVHumanNet may suffer from domain gap. With some time for finetuning, the models outperform the ones trained merely on the train set of HuMMan.}
\vspace{-4mm}
\label{fig:nerf_vis_humman}
\end{figure}

\subsection{Text-driven Image Generation}\label{text_task}
\begin{figure}[tb]
\begin{center}
\includegraphics[width=1.0\linewidth]{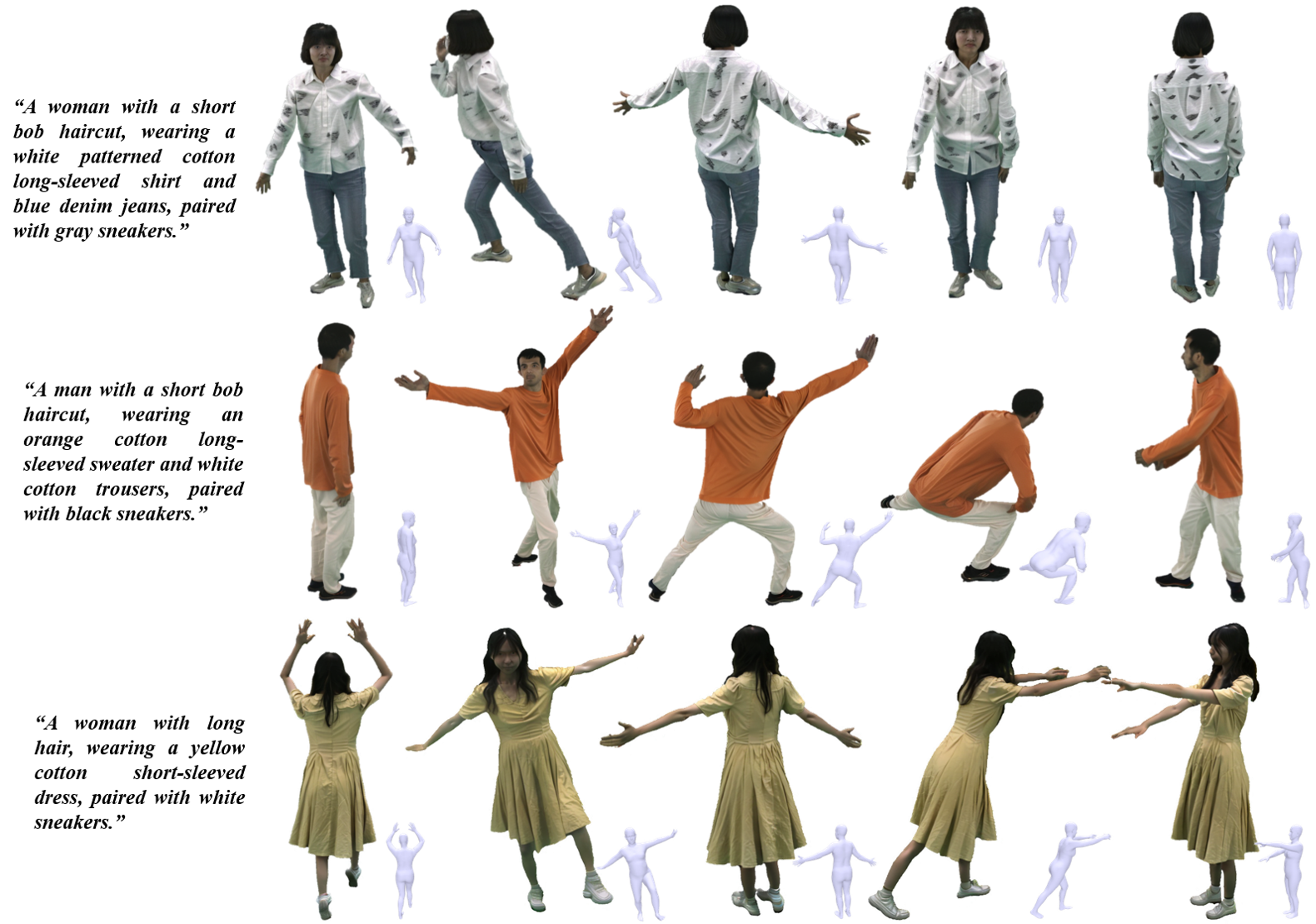}
\end{center}
\vspace{-6mm}
\caption{
\textbf{The visualization of images generated by text-to-image model trained on MVHumanNet with SMPL condition and text prompts as input.} The results demonstrate that training on our large-scale high-quality human dataset enables the generation of high-resolution human images using textual description and SMPL conditions.
Supp. shows more results.}
\label{fig:text_driven image generation}
\vspace{-5mm}
\end{figure}

\thename~is able to serves as a fundamental resource for our text-driven image generation method. The inclusion of comprehensive pose variations within our dataset enhances the potential for generating diverse human images in accordance with text descriptions. 
We finetune the powerful text-to-image model, Stable Diffusion~\cite{rombach2022high} on MVHumanNet dataset to enable text-driven realistic human image generation. 
As shown in \cref{fig:text_driven image generation}, given a text description and a target SMPL pose, we can produce high-quality results with the same consistency as text description and SMPL.


Based on the results derived from the text-driven image generation, it becomes evident that the utilization of large-scale multi-view data from real capture contributes to the efficacy of text-driven realistic human image generation.

\subsection{Human Generative Model }\label{avatar_task}

Recently, generative models have become a prominent and highly researched area. 
Methods such as StyleGAN~\cite{karras2020analyzing, fu2022styleganhuman} have emerged as leading approaches for generating 2D digital human. 
More recently, the introduction of GET3D~\cite{gao2022get3d} has expanded this research area to encompass the realm of 3D generation. 
With the availability of massive data in MVhumanNet, we embark on an exploratory journey as pioneers, aiming to investigate the potential applications of existing 2D and 3D generative models by leveraging a large-scale dataset comprising real-world multi-view full-body data. 
We conduct experiments to unravel the possibilities within this context.

\begin{figure}[tb]
\begin{center}
\includegraphics[width=0.90\linewidth]{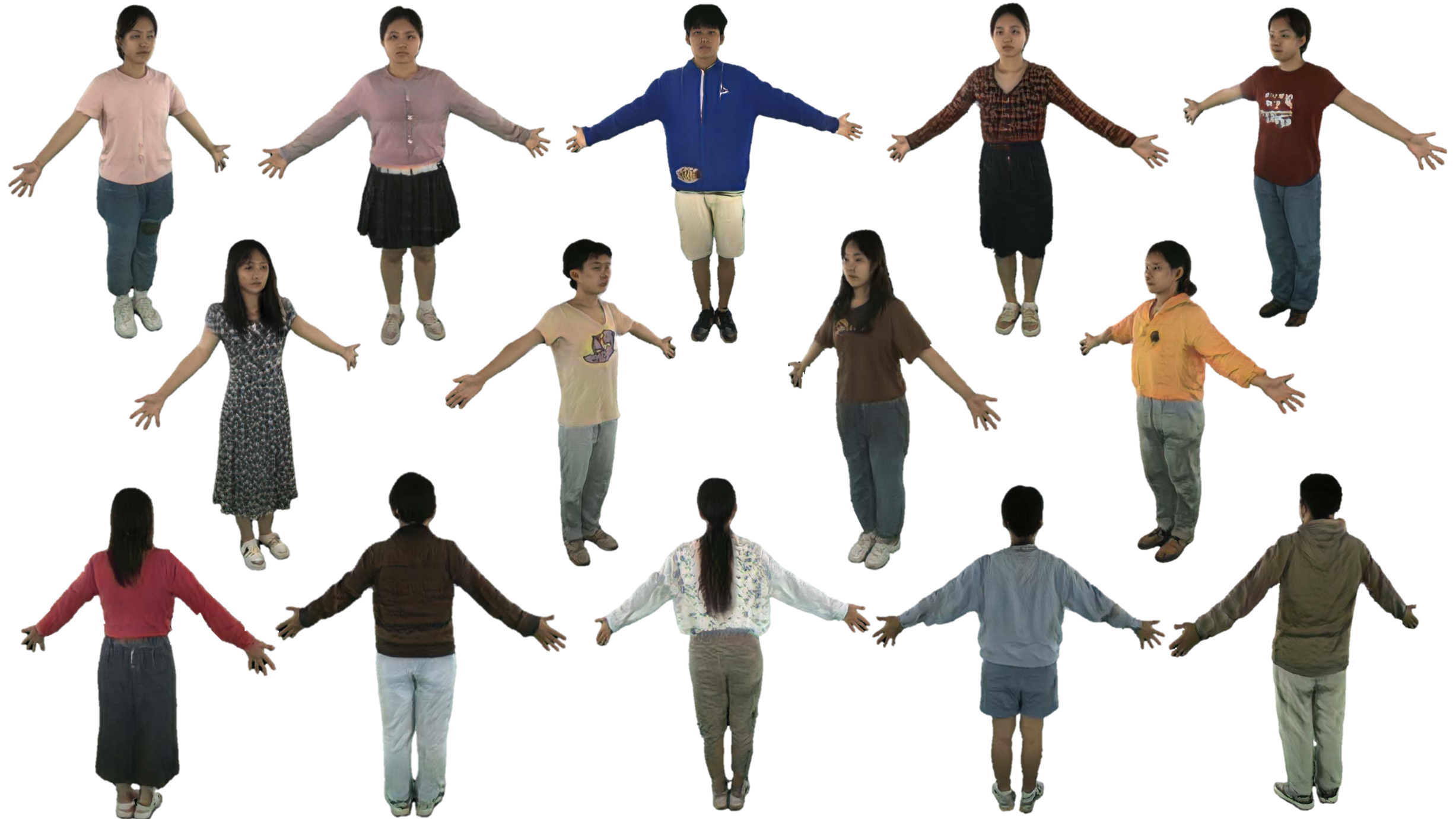}
\end{center}
\vspace{-5mm}
\caption{
\textbf{Visualize the results of StyleGAN2 trained with MVHumanNet.} We randomly sample latent codes from Gaussian distribution and obtain the results. See supp. for more results.}
\label{fig:stylegan2_human}
\vspace{-5mm}
\end{figure}

\noindent\textbf{2D Generative Model} Giving a latent code sampled from Guassian distribution, StyleGAN2 outputs a reasonable 2D images. 
In this part, we feed approximately 198,000 multi-view A-pose images (5500 outfits) and crop to 1024$\times$1024 resolution into the network with camera conditions for training. Fig.~\ref{fig:stylegan2_human}
visualizes the results. 
Our model not only produces frontage full-body images but also demonstrates the capability to generate results from other views, including the back and side views.

\begin{figure}[tb]
\begin{center}
\includegraphics[width=0.99\linewidth]{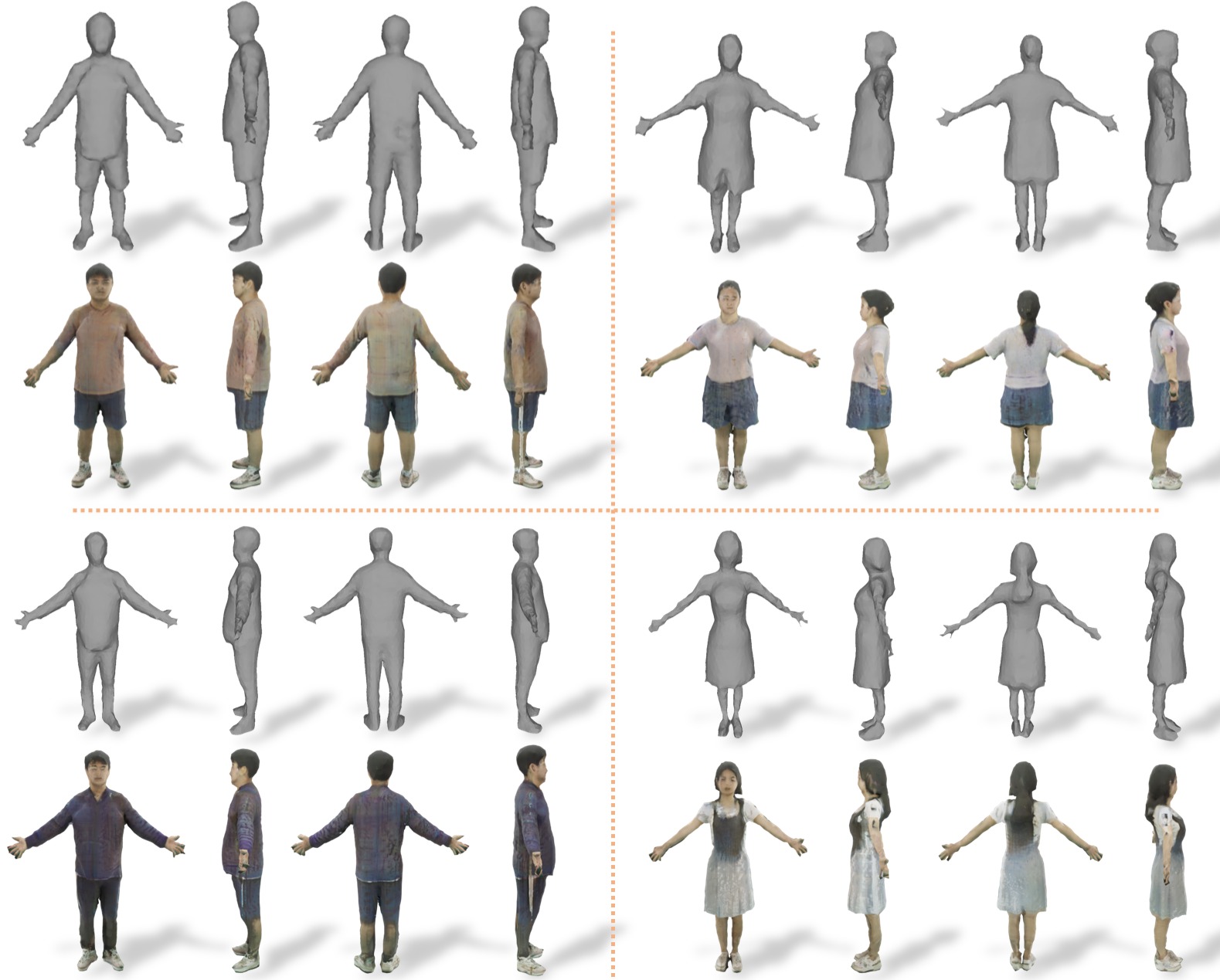}
\end{center}
\vspace{-5mm}
\caption{
\textbf{The visualization results of GET3D trained with MVHumanNet rendered by Blender~\cite{blender}.} The first and third rows represent the geometry, while the second and fourth row shows the texture corresponding to geometry. }
\label{fig:get3d}
\vspace{-2mm}
\end{figure}

\noindent\textbf{3D Generative Model}
Unlike StyleGAN2, GET3D~\cite{gao2022get3d} introduces a distinct requirement of one latent code for geometry and another for texture. 
We use the same amount of data as training StyleGAN2 to train GET3D. 
The visualization results are shown in Fig.~\ref{fig:get3d}. 
The model exhibits the ability to generate reasonable geometry and texture in the A-pose, thereby enabling its application in various downstream tasks.
With the substantial support provided by \thename, various fields, including 3D human generation, can embark on further exploration by transitioning from the use of synthetic data or single-view images to the incorporation of authentic multi-view data.
We also conduct experiments to prove that the performance of the generative model will become more powerful with the increase in the amount of data. The quantitative results are shown in Tab.~\ref{tab:num_generative}. 
We have reason to believe that with the further increase of data, the ability of trained models can further improve.

\begin{table}[tb]
    \centering
    \scalebox{0.8}
    {
    \begin{tabular}{c|cc}
        \toprule
          {\multirow{2}*{Number of Subjects}}  &  \multicolumn{2}{c}{FID$\downarrow$} \\
          & StyleGAN2~\cite{karras2020analyzing}  &  GET3D~\cite{gao2022get3d} \\
        \midrule
          3000 &  14.05 &  41.54   \\
          5500 &  7.08 \textcolor{blue}{(-6.97)} &  25.12 \textcolor{blue}{(-16.42)}\\  
        \bottomrule
    \end{tabular}
    }
    \captionsetup{skip=6pt}
    \caption{\small \textbf{Quantitative comparison of generative models with different data scale}. 
    The performance of both 2D and 3D generative models exhibits obvious improvement with scaling up data.} 
    \label{tab:num_generative}
    \vspace{-3mm}
\end{table}

\section{Conclusion}
In this work, we present MVHumanNet, a large-scale multi-view dataset containing \textbf{4,500} human identities, \textbf{9,000} daily outfits and \textbf{645 million} frames with extensive annotations. We primarily focus on the domain of collecting daily dressing, which allows us to easily scale up the human data. To probe the potential of the proposed large-scale dataset,  we design four experiments  to show how MVHumanNet can be used to power these 3D human-centric tasks.  We plan to release the MVHumanNet dataset with annotations publicly and hope that it will serve as a foundation for further research in the 3D digital human community.

To mitigate potential negative social impacts, we will implement strict regulations on the utilization of our data.


\noindent\textbf{Future Work.}  
We will incorporate all the data to further explore the possibilities of scaling up the training data. In addition, existing human-centric generalizable NeRF methods are designed with significant considerations of data scarcity, which may be adverse to generalization as they highly rely on the coarse representation of human body, i.e. SMPL model. With the largest scale data contained in MVHumanNet, these methods can get rid of the SMPL
model and be redesigned to achieve better generalization. 


\clearpage


{
    \small
   \bibliographystyle{ieeenat_fullname}
 \bibliography{main}
}

\clearpage



\end{document}